\documentclass[a4paper,11pt]{article}

\usepackage{subfig}

\usepackage[symbol]{footmisc}
\renewcommand{\thefootnote}{\fnsymbol{footnote}} 
\usepackage[hidelinks]{hyperref}

\usepackage{fourier}
\usepackage{color}
 \usepackage{graphicx}
\usepackage{url}
\usepackage[affil-it]{authblk}
\usepackage{amsmath}
\usepackage{wrapfig}
\usepackage{graphicx}
\usepackage{multirow}
\usepackage{xcolor,pifont}
\usepackage{amssymb}
\usepackage{fontawesome}
\usepackage{tabularx}
\usepackage{nicematrix}
\usepackage{colortbl}
\usepackage{hhline}
\usepackage{graphicx}
\usepackage{subfig}
\usepackage{csquotes}
\usepackage{caption}
\usepackage[hidelinks]{hyperref}
\usepackage{gensymb}

\usepackage[T1]{fontenc}
\usepackage{times}

\begin{document}

\title{ Navigating Uncertainty: The Role of Short-Term Trajectory Prediction in Autonomous Vehicle Safety}

%\author{Anonymous Submission}
%\affil{Anonymous Affiliation}

\author{Sushil Sharma$^{1,2}$, Ganesh Sistu$^{2}$, Lucie  Yahiaoui$^{2}$, Arindam Das$^{1,3}$, \\ Mark Halton$^{1}$ and Ciar\'{a}n Eising$^{1}$ }
\affil{$^{1}$firstname.lastname@ul.ie and $^{2,3}$firstname.lastname@valeo.com}
\affil{$^{1}$University of Limerick, Ireland, $^{2}$Valeo Vision Systems, Ireland,
$^{3}$DSW, Valeo India}
\date{}
\maketitle
\thispagestyle{empty}

%Turning comments on and off
\ifx00 %ifx01 to turn off and ifx00 to turn on
\newcommand{\AD}[1]{\textcolor{red}{[Arindam: #1]}}
\else 
\newcommand{\AD}[1]{\textcolor{red}{}}
\fi

\newcommand\blfootnote[1]{%
  \begingroup
  \renewcommand\thefootnote{}\footnote{#1}%
  \addtocounter{footnote}{-1}%
  \endgroup
}

\begin{abstract}

Autonomous vehicles require accurate and reliable short-term trajectory predictions for safe and efficient driving. While most commercial automated vehicles currently use state machine-based algorithms for trajectory forecasting, recent efforts have focused on end-to-end data-driven systems. Often, the design of these models is limited by the availability of datasets, which are typically restricted to generic scenarios. To address this limitation, we have developed a synthetic dataset for short-term trajectory prediction tasks using the CARLA simulator. This dataset is extensive and incorporates what is considered complex scenarios - pedestrians crossing the road, vehicles overtaking - and comprises 6000 perspective view images with corresponding IMU and odometry information for each frame. Furthermore, an end-to-end short-term trajectory prediction model using convolutional neural networks (CNN) and long short-term memory (LSTM) networks has also been developed. This model can handle corner cases, such as slowing down near zebra crossings and stopping when pedestrians cross the road, without the need for explicit encoding of the surrounding environment. In an effort to accelerate this research and assist others, we are releasing our dataset and model to the research community. Our datasets are publicly available on \textcolor{purple}{\href{https://github.com/sharmasushil/Navigating-Uncertainty-Trajectory-Prediction}{https://github.com/sharmasushil/Navigating-Uncertainty-Trajectory-Prediction}}. 
\end{abstract}


%%%%%%%%%%%%%%%%%%%%%%
\section{Introduction}

Autonomous vehicles are revolutionizing the transportation industry with a core focus on improved safety and an enhanced driver experience. At the same time, ensuring the safe manoeuvring of autonomous vehicles in real-world scenarios remains very challenging. One of the key components of autonomous vehicle safety is the ability to accurately predict short-term trajectories that allow the host vehicle to navigate through uncertain and dynamic environments \cite{zhao2020enhanced}. Short-term trajectory prediction refers to the estimation of the future position and movement within a limited time frame of a vehicle. By accurately perceiving the ego vehicle trajectory, an autonomous vehicle can anticipate potential hazards and proactively plan its actions to avoid collisions \cite{vehicles3030036} or respond to risky situations\cite{BOTELLO2019100012}. Significant progress has been made in recent years in developing trajectory prediction models for autonomous vehicles \cite{venkatesh2023connected}. To make predictions about the future movements of surrounding entities, these models employ diverse data sources, including sensor data like LiDAR, radar, and cameras. Machine learning techniques, including deep neural networks \cite{s20154220}, have proven to be effective in capturing complex spatiotemporal patterns \cite{chen2023traj} and improving trajectory prediction accuracy. By analyzing and understanding previous data, researchers and engineers can identify common patterns, critical factors, and potential risks associated with trajectory prediction \cite{cui2019multimodal}. This knowledge can guide the development of more reliable and effective prediction algorithms, leading to enhanced safety measures \cite{atakishiyev2021explainable} and increased public trust in autonomous vehicles. In this study, we aim to investigate the role of short-term trajectory prediction in ensuring the safety of autonomous vehicles. 
The main contributions of this  paper are as follows:

\begin{enumerate}
\item Short-term trajectory prediction of the vehicle from only perspective view images with no explicit knowledge encoding.
\item A novel dataset\footnote{\textcolor{purple}{\href{https://drive.google.com/drive/folders/1JPb64bGV88ymZkJrUBaKQg12tToZVF7T?usp=sharing}{Dataset: https://drive.google.com/drive/folders/1JPb64bGV88ymZkJrUBaKQg12tToZVF7T?usp=sharing}}} to encourage the research community to pursue the direction of end-to-end implicit vehicle trajectory prediction learning methods. 
\end{enumerate}

\section{Related Work}

Several studies have focused on the crucial role of short-term trajectory prediction in ensuring the safety of autonomous vehicles\cite{9294553}. One notable work is the research conducted by \cite{zhu2019probabilistic}. The authors proposed the use of Recurrent Neural Networks (RNNs) to accurately forecast the future trajectories of surrounding objects in complex driving environments. By training their model on real-world driving datasets, they achieved impressive trajectory prediction results, allowing autonomous vehicles to navigate with improved safety and awareness. Another relevant study by \cite{9243464}, explored the application of Generative Adversarial Networks (GANs) for probabilistic trajectory prediction. By leveraging GANs, the researchers were able to generate multiple plausible future trajectories for vehicles, incorporating uncertainty into the prediction process. This probabilistic approach provides valuable information for autonomous vehicles to assess potential risks and make informed decisions in dynamic traffic situations. Furthermore, the work of \cite{li2019grip++}, emphasized the significance of considering interaction information between vehicles and pedestrians. The researchers proposed a novel interaction-aware trajectory prediction model \cite{kruger2020interaction} that effectively captured the mutual influence and dependencies between different road users. By incorporating interaction information, the model achieved improved accuracy \cite{deo2018convolutional} and reliability in trajectory prediction, contributing to enhanced safety in autonomous driving scenarios. These studies collectively highlight the importance of short-term trajectory prediction in autonomous vehicle safety. Our study aims to address the limitations encountered in previous studies by incorporating frequently occurring critical actors - pedestrians at the crosswalks, keeping the volume of the dataset minimal and balanced while ensuring all critical cases are covered. To address these challenges, we have created a novel dataset using the Carla simulation platform. By leveraging this synthetic dataset, we expect our model to achieve enhanced performance. This approach enables us to overcome the constraints related to data collection and processing, thereby augmenting the results of our study.

\section{Methodology}
\subsection{Architecture Topology}

The network developed in this work is designed to predict the future trajectories of a vehicle based on a sequence of perspective-view images. It comprises two primary components: a Convolutional Neural Network (CNN) and a Long-Short Term Memory Network (LSTM). CNN plays a crucial role in extracting essential features from the input image sequence using convolution, a mathematical operation that filters the data to capture specific features. These deep features obtained from the CNN serve as inputs to the LSTM, which specializes in temporal prediction tasks by capturing long-term dependencies and maintaining memory over time. The LSTM network learns to infer future positions of the vehicle within the predicted trajectory based on the extracted deep features and the input image sequence.

\begin{figure}[h!]
\centering
\includegraphics[width=\columnwidth]{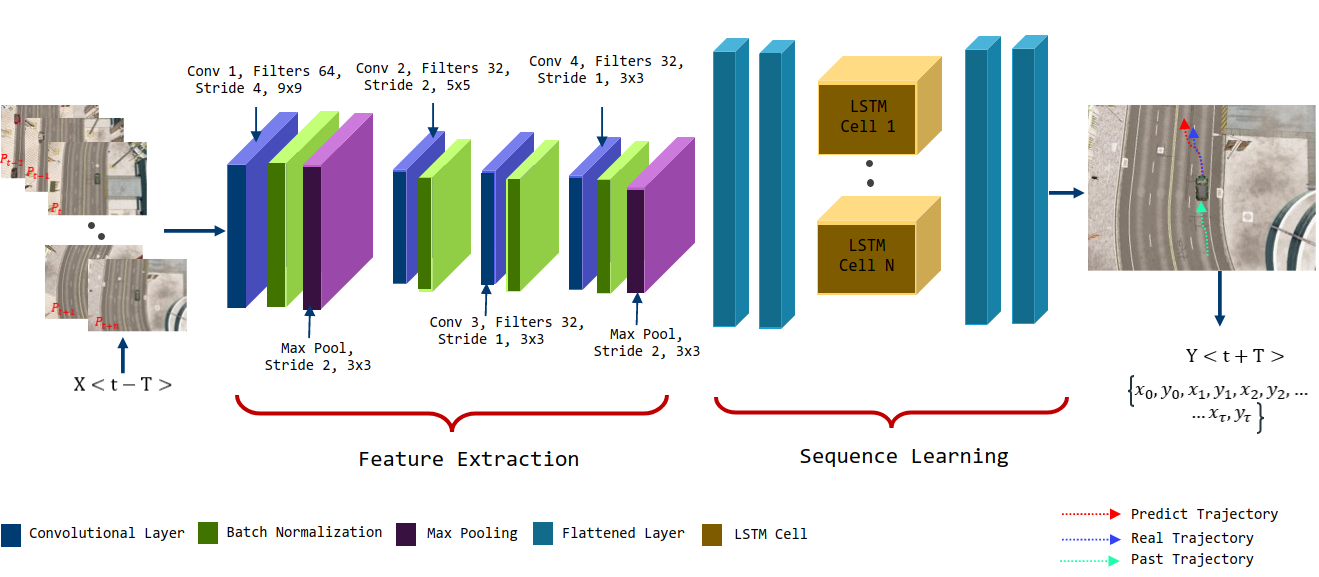}
\caption{Proposed CNN architecture topology diagram for trajectory estimation from image sequences, showcasing the flow of information through convolutional and fully connected layers.} %\AD{Please add a detailed caption}
\label{fig:boschVirtualVisor}
\end{figure}

Overall, the architecture of the network is shown in Figure \ref{fig:boschVirtualVisor}, which provides a visual summary of how the CNN and LSTM components are connected. The architecture topology diagram illustrates the CNN used for trajectory estimation from a sequence of $n$ input images. The left side of the diagram shows the input layer of the network where the image sequence is fed into the CNN. The images are then processed through a series of convolutional layers, pooling layers, and activation functions to extract features from the images. The extracted features are then passed through one or more fully connected layers to estimate the trajectory, which is displayed on the right side of the diagram. We use a custom encoder here since this work aims to provide a solution that can be deployed on a low-resource device within less than 1 TOPS. However, given no constraints on the device, any state-of-the-art encoder as a backbone can be integrated into the feature extraction stage.

\subsection{Trajectory prediction approach }

We define a sequence as a sequence of \textit{ n}  images as \begin{math}X^{<t-n>}\end{math}  at time t. The purpose of \begin{math}X^{<t-n>}\end{math} is to anticipate future trajectory position.
\begin{equation}
Y^{<t+n>} = \bigl\{x_0,y_0,x_1,y_1,x_2,y_2,x_3,y_3,. . . . . .,x_n,y_n\bigl\}
\end{equation}
where \textit{x} and \textit{y} depict the location of the ego vehicle. The distance feedback \begin{math}d_1^{<t-n>}\end{math} measures the distance between the current pose of the ego vehicle $P_{ego}$ and the final pose in the sequence $P_{dest}$ and expressed in (\ref{eq:mylabel1}). This objective's goal is to shorten the ego vehicle's local travel path.
\begin{equation}
d_1^{<t+n>} = \sum_{i=1} ^{n_0} \Big\| P_{ego}^{<t+i>} - P_{dest}^{<t+n>} \Big\|_2^2 \label{eq:mylabel1}
\end{equation}
(\ref{eq:mylabel2}) demonstrates how to derive the lateral velocity $V_{lat}$ from the angular velocity of the ego vehicle $v_\delta$. The main objective in terms of trajectory prediction is to anticipate the future path of the ego vehicle based on its current state and inputs. By understanding the vehicle's lateral velocity $V_{lat}$ and angular velocity $v_\delta$, we can estimate its trajectory and predict its future position and orientation.

\begin{equation}
V_{lat}^{<t+n>} = \sum_{i=1} ^{n_0} v_\delta^{<t+n>} \label{eq:mylabel2}
\end{equation}
(\ref{eq:mylabel3}) defines the $y$ direction velocity component, which is used to determine the longitudinal velocity as $V_{long}$. The speed is set at 30 kph.  The main objective is to calculate the future longitudinal velocity denoted as $V_{long}^{<t+n>}$, based on the sum of forward velocities, represented as $v_f^{<t+n>}$, over a specific time horizon.
\begin{equation}
V_{long}^{<t+n>} = \sum_{i=1} ^{n_0} v_f^{<t+n>} \label{eq:mylabel3}
\end{equation}
Root mean squared error (RMSE) is also being tested as an objective, which is represented by the Euclidean distance between the predicted position of the ego vehicle $\widehat{P}_{ego}$ for a given trajectory at a specific time-step and the actual position of the ego vehicle $P_{ego}$ at that time-step, as seen in equation (\ref{eq:mylabel4}).

\begin{equation}
    RMSE = \sum_{i=1} ^{n} \frac{\sqrt{(\widehat{P}_{ego} - P_{ego})^2}}{n} \label{eq:mylabel4}
\end{equation}

\subsection{Why the CARLA simulation platform?}

We have chosen to utilize the CARLA simulation platform \cite{Dosovitskiy17} for creating our dataset for a number of reasons. It is a dedicated open-source simulator for autonomous driving research. When compared to other simulation platforms such as Carsim \cite{dupuy2001generating}, MATLAB \cite{herrera2016gogps}, and Gazebo \cite{10.3389/frobt.2021.713083}, the CARLA simulation stands out for its comprehensive feature set, realistic simulation, customization options, and support for multi-sensor data generation. 

A comparison of each of the aforementioned simulation platforms is given in Table \ref{tab:varioussimulators}.

 \newcommand*\colourcheck[1]{%
  \expandafter\newcommand\csname #1check\endcsname{\textcolor{#1}{\ding{52}}}%
}
%\colourcheck{green}
\colourcheck{blue}

\begin{table*}[h!]
\centering

\scalebox{0.8}{
\begin{tabular}{|l|l|l|l|l|}
\hline
\multicolumn{5}{|c|}{\textbf{Simulation}} \\
\hline
\textbf{Requirement} & \textbf{CARLA} & \textbf{CarSim} &\textbf{ MATLAB} & \textbf{Gazebo}\\
\hline
\textbf{Perception: }sensor models supported &  \hspace{4.5mm}\bluecheck  &  \hspace{4.5mm}\bluecheck   &  \hspace{4.5mm}\bluecheck   &  \hspace{4.5mm}\bluecheck  \\ \hline
\textbf{Perception:} support for various weather conditions &   \hspace{4.5mm}\bluecheck   &  \hspace{4.5mm} \color{red} \faTimes  &  \hspace{4mm} \color{red} \faTimes &  \hspace{4mm} \color{red} \faTimes \\ \hline
\textbf{Camera Calibration} &  \hspace{4.5mm}\bluecheck   &  \hspace{4.5mm} \color{red} \faTimes  &  \hspace{4.5mm}\bluecheck   &  \hspace{4.5mm}\bluecheck  \\ \hline
\textbf{3D Virtual Environment}  &   \hspace{4.5mm}\bluecheck   & \hspace{4.5mm}\bluecheck   &  \hspace{4.5mm} \color{red} \faTimes &  \hspace{4.5mm}\bluecheck  \\ \hline
\textbf{Path planing} &   \hspace{4.5mm}\bluecheck   &  \hspace{4.5mm}\bluecheck    &  \hspace{4.5mm}\bluecheck   &  \hspace{4.5mm}\bluecheck  \\ \hline
\textbf{Traffic scenarios simulation} &  \hspace{4.5mm}\bluecheck  & \hspace{4.5mm}\bluecheck  &  \hspace{4.5mm}\bluecheck   &  \hspace{4.5mm} \color{red} \faTimes \\ \hline
\textbf{Scalability }via a server multi-client architecture  &   \hspace{4.5mm}\bluecheck   &  \hspace{4.5mm} \color{red} \faTimes & \hspace{4.5mm} \color{red} \faTimes  & \hspace{4.5mm} \color{red} \faTimes \\ \hline
\textbf{Portability} such as window and Linux &   \hspace{4.5mm}\bluecheck   & \hspace{4.5mm}\bluecheck   &  \hspace{4.5mm}\bluecheck   &  \hspace{4.5mm}\bluecheck  \\ \hline
\textbf{Flexible API}  &   \hspace{4.5mm}\bluecheck   &  \hspace{4.5mm}\bluecheck    &  \hspace{4.5mm}\bluecheck   &  \hspace{4.5mm}\bluecheck  \\ \hline
\textbf{Open Source} &  \hspace{4.5mm}\bluecheck   & \hspace{4.5mm} \color{red} \faTimes  &  \hspace{4.5mm} \color{red} \faTimes  &  \hspace{4.5mm}\bluecheck   \\
\hline
\end{tabular}
}
\caption{Comparison of various simulators.}
\label{tab:varioussimulators}

\end{table*}

\section {Experimental Setup}
\subsection{Implementation Details}

A CNN-LSTM model for trajectory prediction is implemented using the TensorFlow framework and trained on perspective view images from the CARLA simulation. The first step is preprocessing, which involves resizing, normalizing, and splitting the dataset into a ratio of 60\% : 20\%: 20\% for training, validation and testing respectively. The next step involves a CNN extracting spatial features, which are then fed into an LSTM to model for temporal dependencies. The training process involves utilizing suitable loss functions to train the model effectively. In the case of trajectory prediction, one commonly used loss function is the Mean Squared Error (MSE) loss. This particular loss function calculates the average of the squared differences between the predicted trajectories and the ground truth trajectories. By doing so, it penalizes larger deviations between the predicted and actual trajectories, we specifically employed the Adam optimization algorithm \cite{liu2023improved}.  
Table \ref{tab:hypermeter1} outlines the hyperparameters employed to train the proposed CNN-LSTM networks. The optimal values of these hyperparameters are obtained 
experimentally to maximize the performance and increase the generalization capabilities of the model.

\begin{table}[htbp]
\centering
\setlength{\arrayrulewidth}{0.8pt}
  \begin{center}
    \label{tab:table1}
    \scalebox{0.8}{
    \begin{tabular}{lll} 
    \hline
      \textbf{S.No }\,\,\,\,\,\,\,\,\,\,\,\,\,\  & \,\,\,\,\,\,\,\,\,\,\,\,\,\ \textbf{Parameter Name } & \,\,\,\,\,\,\,\,\,\,\,\,\ \textbf{Optimal Values}\\
      \hline
      \hline
      1\,\,\,\,\,\,\,\,\,\,\,\,\ & \,\,\,\,\,\,\,\,\,\,\,\, Batch Size\,\,\,\,\,\,\,\,\,\,\,\,\ & \,\,\,\,\,\,\,\,\,\,\,\,\ ${\big(50,75,100\big)}$ \\
      2 & \,\,\,\,\,\,\,\,\,\,\,\,\ Epochs & \,\,\,\,\,\,\,\,\,\,\,\,\ {\big(10,20,30,40\big)} \\
      3 &\,\,\,\,\,\,\,\,\,\,\,\,\ Loss Function\,\,\,\,\,\,\,\,\,\,\,\,\ & \,\,\,\,\,\,\,\,\,\,\,\,\ \big(Mean square error (MSE)\big) \\
       4 &\,\,\,\,\,\,\,\,\,\,\,\,\ Momentum \,\,\,\,\,\,\,\,\,\,\,\,\ & \,\,\,\,\,\,\,\,\,\,\,\,\ \big(0.8,0.85,0.9\big) \\
       5 &\,\,\,\,\,\,\,\,\,\,\,\,\ Optimizer \,\,\,\,\,\,\,\,\,\,\,\,\ & \,\,\,\,\,\,\,\,\,\,\,\,\ \big(Nadam, Adam\big) \\
        6 &\,\,\,\,\,\,\,\,\,\,\,\,\ LSTM cells  \,\,\,\,\,\,\,\,\,\,\,\,\ & 
        \,\,\,\,\,\,\,\,\,\,\,\,\  \big(1,2,3,4\big) \\
      7 &\,\,\,\,\,\,\,\,\,\,\,\,\ LSTM Dropout \,\,\,\,\,\,\,\,\,\,\,\,\ & \,\,\,\,\,\,\,\,\,\,\,\,\ \big(0.25,0,3,0.35,0.4\big) \\
       8 &\,\,\,\,\,\,\,\,\,\,\,\,\ Hidden Units  \,\,\,\,\,\,\,\,\,\,\,\,\ & \,\,\,\,\,\,\,\,\,\,\,\,\ \big(100,125,150,175,200\big) \\
       9 &\,\,\,\,\,\,\,\,\,\,\,\,\ CNN Flattened 1  \,\,\,\,\,\,\,\,\,\,\,\,\ & \,\,\,\,\,\,\,\,\,\,\,\,\ \big(256,512,768,1024\big) \\
       10 &\,\,\,\,\,\,\,\,\,\,\,\,\ CNN Flattened 2  \,\,\,\,\,\,\,\,\,\,\,\,\ & \,\,\,\,\,\,\,\,\,\,\,\,\ \big(256,512,768,1024\big) \\
        11 &\,\,\,\,\,\,\,\,\,\,\,\,\ LSTM Flattened 1  \,\,\,\,\,\,\,\,\,\,\,\,\ & \,\,\,\,\,\,\,\,\,\,\,\,\ \big(64,128,256,512\big) \\
         12 &\,\,\,\,\,\,\,\,\,\,\,\,\ LSTM Flattened 2  \,\,\,\,\,\,\,\,\,\,\,\,\ & \,\,\,\,\,\,\,\,\,\,\,\,\ \big(64,128,256,512\big) \\
         13 &\,\,\,\,\,\,\,\,\,\,\,\,\ Flattened Dropout  \,\,\,\,\,\,\,\,\,\,\,\,\ & \,\,\,\,\,\,\,\,\,\,\,\,\ \big(0.05,0.1,0.15,0.2.0.25\big) \\
         \hline
         \hline

    \end{tabular}
    }
    \caption{The table provides a list of hyperparameters that include batch size, loss function, optimiser, and others for our proposed training setup.} 
    \label{tab:hypermeter1}
  \end{center}
  
\end{table}

\vspace{10mm}
\subsection{Dataset Generation}

The specifics of how the datasets were created using the CARLA simulator are described in this section. In the simulator, we adjusted the camera position to achieve a top-down view, enabling us to capture comprehensive 360\degree and Bird's Eye View (BEV) perspectives of each scene. Note that the utilization of the top-view image approach for trajectory prediction in autonomous vehicles provides a comprehensive and detailed comprehension of the surrounding environment, which in turn aids in making accurate decisions.  Each image has dimensions of 800 pixels width and 600 pixels height. The field of view (FOV) for the camera is set to 90$\degree$. The CARLA camera position and orientation are defined as cam\_rotation = (-90\textdegree, 0\textdegree, -90\textdegree) and cam\_location = $\mathsf{(0,0,15)}$. If we consider a camera positioned 15 meters above the host vehicle, it would imply that the camera is mounted at a height of 15 meters from the ground level. This positioning suggests that the camera is elevated significantly above the vehicle, capturing a top-down or bird's eye view perspective of the surrounding environment.

\begin{figure}[h!]
    \centering
    \subfloat[\centering Dataset: Level 1 with 1000 images - Illustrates the distribution of objects in the scene categorized by class.]{{\includegraphics[width=8cm]{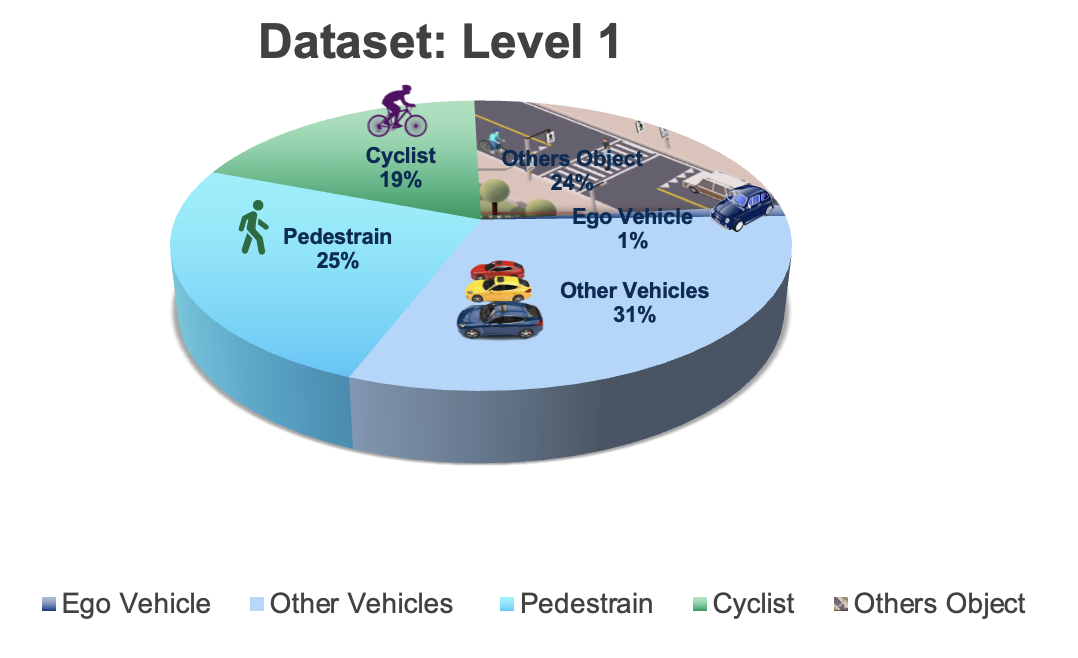} }}%
    \qquad
    \subfloat[\centering Dataset: Level 2 with 5000 images - Depicts the class-wise object representation within the scene.]{{\includegraphics[width=8cm]{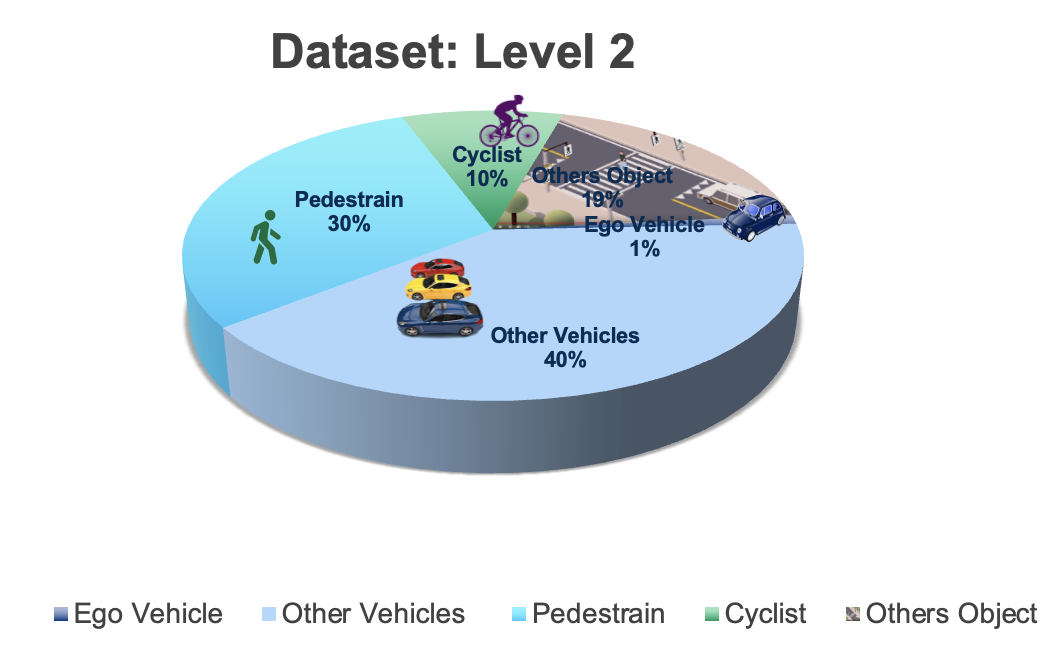} }}%
    \caption{Data statistics for each class of our newly created dataset}%
    \vspace{-0.5cm}%
    \label{fig:example}%
\end{figure}

The initial dataset, referred to as ``Level 1'', consists of 1000 perspective view images. To achieve a more enhanced real-life simulation, ego vehicles and pedestrians were incorporated into each scene. In addition, we annotated each image with supplementary details such as speed and local coordinates (x,y and z).
Similarly, the subsequent, more challenging dataset, referred to as ``Level 2'', consists of 5000 images. In order to generate more intricate and diverse scenarios, the number of vehicles and pedestrians was augmented in each scene. This deliberate augmentation aimed to ensure that our machine-learning models could effectively handle a wide range of real-life scenarios. As done previously, all images in this dataset were annotated with supplementary information, as presented in Figure \ref{fig:example}.

\vspace{-6mm}

\section{Results}
In the Carla simulation, we define an IMU sensor that captures linear acceleration (m/s$^2$) and angular velocity data (rad/sec). The IMU sensor records the frame number at which each measurement occurs, enabling us to determine the simulation start time accurately. By utilizing a sensor, we can collect real-time measurements of linear acceleration and angular velocity throughout the simulation. The time elapsed from when the simulation started to when the measurement was taken was also recorded. The GNSS sensor provides the position and rotation of the sensor at the time of measurement, relative to the vehicle coordinates. The position is typically represented in meters, while the rotation is expressed in degrees. A comparison between the ground truth trajectory and our predicted trajectory is shown in Figure 3(a). In Figure 3(b), we focus on a specific critical scenario where a pedestrian enters the road, leading to vehicles coming to a halt. For more insight, a demo video on this real-life scenario can be checked in detail at \textcolor{purple}{\href{https://youtu.be/DZDqGbkInko?t=31}{https://youtu.be/DZDqGbkInko?t=31}}

An ablation study on the number of LSTM cells ($\alpha$=1, $\beta$=2, $\gamma$=3, $\delta$=4) is conducted on our CNN-LSTM model. This comparison was performed using the CARLA dataset for the two specified levels, Level 1 and Level 2 respectively. For this analysis, three evaluation metrics - RMSE, MAPE, and AED. A summary of the results is shown in Table \ref{table:3}. 

\begin{figure}[h!]
    \centering
    \subfloat[\centering Tracking the Path: Ground Truth vs. Predicted Trajectory ]{{\includegraphics[width=8.1cm]{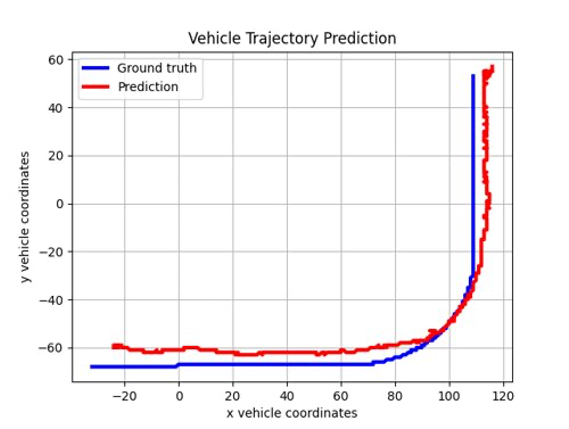} }}%
    \qquad
    \subfloat[\centering Unleashing Velocity: Vehicles velocity  at 30 km/h]{{\includegraphics[width=7.5cm]{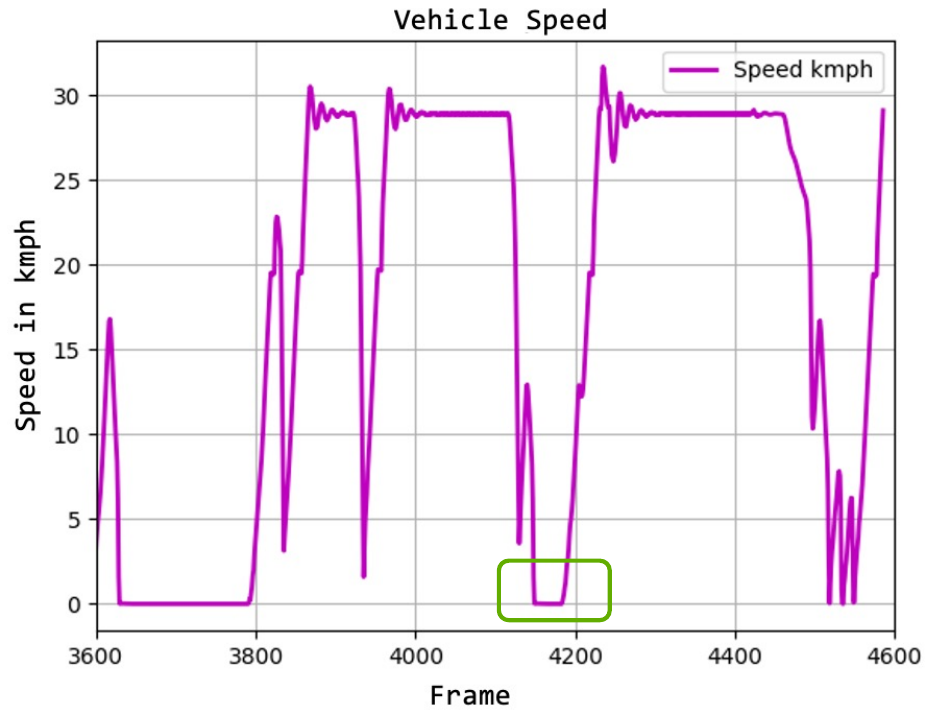} }}%
    \caption{CARLA simulation results  }%
    \label{fig:example56}%
    \vspace{-0.2cm}%
\end{figure}

\begin{table*}[h]
\setlength{\arrayrulewidth}{1pt}

\centering
\scalebox{0.8}{
\begin{tabular}{ p{3.0cm} | p{2.9cm} | p{1.8cm} | p{1.5cm} | p{1.5cm} p{1.5cm} p{1.5cm} } 

\hline
\multicolumn{4}{c}{} \\
\textbf{CARLA: Dataset} &  Model & ARMSE & AMAPE  & AED \\
\multicolumn{4}{c}{} \\
\hline
\textbf{Dataset:} Level 1 & \rotatebox{0}{CNN-LSTM ($\alpha$)}   \rotatebox{0}{CNN-LSTM ($\beta$)}  \rotatebox{0}{CNN-LSTM ($\gamma$)} \textbf{\rotatebox{0}{CNN-LSTM ($\delta$)}}   &  \rotatebox{0}{0.0046}  \rotatebox{0}{0.0034}  \rotatebox{0}{0.0028} \textbf{\rotatebox{0}{0.0024}}&  \rotatebox{0}{0.0056}  \rotatebox{0}{0.0043}  {\rotatebox{0}{0.0038}} \textbf{\rotatebox{0}{0.0033}} &  \rotatebox{0}{0.0050}  \rotatebox{0}{0.0039}  \rotatebox{0}{0.0032}  \textbf{\rotatebox{0}{0.0028}}   \\ 
\hline

\textbf{Dataset:} Level 2 & \rotatebox{0}{CNN-LSTM ($\alpha$)}  \rotatebox{0}{CNN-LSTM ($\beta$)}   \rotatebox{0}{CNN-LSTM ($\gamma$)}  \textbf{\rotatebox{0}{CNN-LSTM ($\delta$)}}&  \rotatebox{0}{0.0126}  \rotatebox{0}{0.0097}  \rotatebox{0}{0.0082} \textbf{\rotatebox{0}{0.0065}}  &
 \rotatebox{0}{0.0172}  \rotatebox{0}{0.0133}  \rotatebox{0}{0.0119} \textbf{\rotatebox{0}{0.0107}} &  \rotatebox{0}{0.0154}  \rotatebox{0}{0.0127}  \rotatebox{0}{0.0107} \textbf{\rotatebox{0}{0.0079} }\\

\hline

\end{tabular}
}
\caption{Analysis of ARMSE, AMAPE, and AED}
\label{table:3}
\end{table*}

\begin{figure*}[t]
  \centering
  \includegraphics[width=0.8\linewidth]{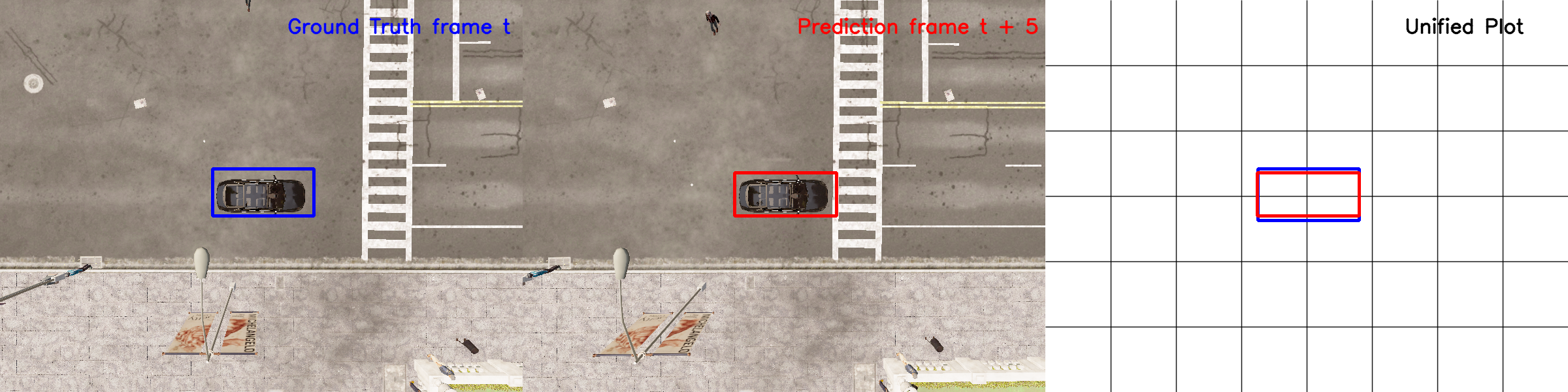}  
  \includegraphics[width=0.8\linewidth]{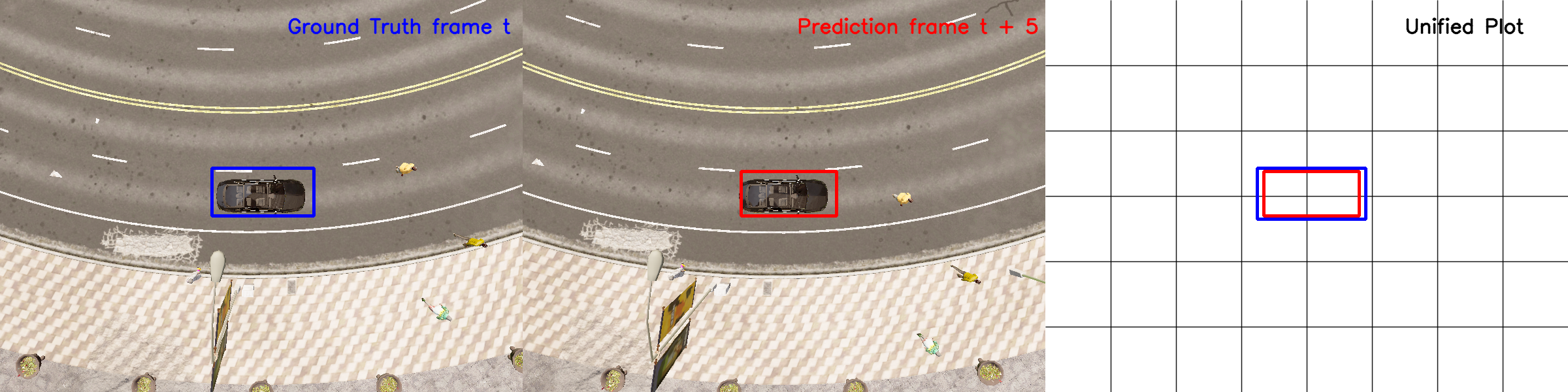}  
\caption{Demonstrating robustness of our model in handling corner cases, such as pedestrian crossings, without explicit encoding of surrounding knowledge. The comparison of Ground Truth \textcolor{blue}{(blue)} and Predicted Bounding Boxes \textcolor{red}{(red)} for trajectory prediction} 

\label{fig:fig}
\end{figure*}

One corner case is presented in Figure \ref{fig:fig}. For visual purposes, a blue bounding box represents the position of the object as per the ground truth and a red bounding box is used to highlight the prediction of the same object from our proposed model. Moreover, our model demonstrates exceptional performance in predicting the frame at time $t+5$, closely aligning with the ground truth $t$. Notably, it excels even in challenging scenarios, such as when a vehicle is navigating a bend and a pedestrian unexpectedly appears. The model effectively captures and comprehends the unpredictable behaviour of pedestrians when crossing the road, showcasing its remarkable trajectory prediction capabilities. Our model performance in these critical scenarios is demonstrated at \textcolor{purple}{\href{https://youtu.be/DZDqGbkInko}{https://youtu.be/DZDqGbkInko}}

\section{Conclusions}

In this work, we have developed a novel single-stage end-to-end deep network proposal for short-term vehicle trajectory prediction.
First, we introduce a CNN-LSTM network topology for trajectory prediction, leveraging its effectiveness in handling complex stochastic tasks. 
Then we generate a large synthetic dataset using the CARLA simulator, providing a valuable resource for training and evaluating trajectory prediction models in a supervised learning fashion with a focus on safety. The data-driven approach presented in this paper offers a scalable alternative to traditional rule-based optimization algorithms, paving the way for further advancements in the field. The provided synthetic dataset serves as a baseline for future research, encouraging the research community to compare their models against the proposed methodology. We hope this effort will foster innovation and drive improvements in trajectory prediction models.

\section*{Acknowledgments}

This article has emanated from research conducted with the financial support of Science Foundation Ireland under Grant number 18/CRT/6049.

\appendix

\bibliographystyle{apalike}

\bibliography{imvip}

\begin{thebibliography}{}

\bibitem[Atakishiyev et~al., 2021]{atakishiyev2021explainable}
Atakishiyev, S., Salameh, M., Yao, H., and Goebel, R. (2021).
\newblock Explainable artificial intelligence for autonomous driving: a
  comprehensive overview and field guide for future research directions.
\newblock {\em arXiv preprint arXiv:2112.11561}.

\bibitem[Botello et~al., 2019]{BOTELLO2019100012}
Botello, B., Buehler, R., Hankey, S., Mondschein, A., and Jiang, Z. (2019).
\newblock Planning for walking and cycling in an autonomous-vehicle future.
\newblock {\em Transportation Research Interdisciplinary Perspectives},
  1:100012.

\bibitem[Chen et~al., 2023]{chen2023traj}
Chen, H., Wang, J., Shao, K., Liu, F., Hao, J., Guan, C., Chen, G., and Heng,
  P.-A. (2023).
\newblock Traj-mae: Masked autoencoders for trajectory prediction.
\newblock {\em arXiv preprint arXiv:2303.06697}.

\bibitem[Cui et~al., 2019]{cui2019multimodal}
Cui, H., Radosavljevic, V., Chou, F.-C., Lin, T.-H., Nguyen, T., Huang, T.-K.,
  Schneider, J., and Djuric, N. (2019).
\newblock Multimodal trajectory predictions for autonomous driving using deep
  convolutional networks.
\newblock In {\em 2019 International Conference on Robotics and Automation
  (ICRA)}, pages 2090--2096. IEEE.

\bibitem[Deo and Trivedi, 2018]{deo2018convolutional}
Deo, N. and Trivedi, M.~M. (2018).
\newblock Convolutional social pooling for vehicle trajectory prediction.
\newblock In {\em Proceedings of the IEEE conference on computer vision and
  pattern recognition workshops}.

\bibitem[Dixit et~al., 2021]{vehicles3030036}
Dixit, A., Kumar~Chidambaram, R., and Allam, Z. (2021).
\newblock Safety and risk analysis of autonomous vehicles using computer vision
  and neural networks.
\newblock {\em Vehicles}, 3(3):595--617.

\bibitem[Dosovitskiy et~al., 2017]{Dosovitskiy17}
Dosovitskiy, A., Ros, G., Codevilla, F., Lopez, A., and Koltun, V. (2017).
\newblock {CARLA}: {An} open urban driving simulator.
\newblock In {\em Proceedings of the 1st Annual Conference on Robot Learning}.

\bibitem[Dupuy et~al., 2001]{dupuy2001generating}
Dupuy, S., Egges, A., Legendre, V., and Nugues, P. (2001).
\newblock Generating a 3d simulation of a car accident from a written
  description in natural language: The carsim system.
\newblock {\em arXiv preprint cs/0105023}.

\bibitem[Fayyad et~al., 2020]{s20154220}
Fayyad, J., Jaradat, M.~A., Gruyer, D., and Najjaran, H. (2020).
\newblock Deep learning sensor fusion for autonomous vehicle perception and
  localization: A review.
\newblock {\em Sensors}, 20(15).

\bibitem[Hegde et~al., 2020]{9243464}
Hegde, C., Dash, S., and Agarwal, P. (2020).
\newblock Vehicle trajectory prediction using gan.
\newblock In {\em 2020 Fourth International Conference on I-SMAC (IoT in
  Social, Mobile, Analytics and Cloud) (I-SMAC)}.

\bibitem[Herrera et~al., 2016]{herrera2016gogps}
Herrera, A.~M., Suhandri, H.~F., Realini, E., Reguzzoni, M., and de~Lacy, M.~C.
  (2016).
\newblock gogps: open-source matlab software.
\newblock {\em GPS solutions}, 20:595--603.

\bibitem[Kr{\"u}ger et~al., 2020]{kruger2020interaction}
Kr{\"u}ger, M., Novo, A.~S., Nattermann, T., and Bertram, T. (2020).
\newblock Interaction-aware trajectory prediction based on a 3d spatio-temporal
  tensor representation using convolutional--recurrent neural networks.
\newblock In {\em 2020 IEEE Intelligent Vehicles Symposium (IV)}, pages
  1122--1127. IEEE.

\bibitem[Li et~al., 2019]{li2019grip++}
Li, X., Ying, X., and Chuah, M.~C. (2019).
\newblock Grip++: Enhanced graph-based interaction-aware trajectory prediction
  for autonomous driving.
\newblock {\em arXiv preprint arXiv:1907.07792}.

\bibitem[Liu et~al., 2023]{liu2023improved}
Liu, M., Yao, D., Liu, Z., Guo, J., Chen, J., et~al. (2023).
\newblock An improved adam optimization algorithm combining adaptive
  coefficients and composite gradients based on randomized block coordinate
  descent.
\newblock {\em Computational Intelligence and Neuroscience}, 2023.

\bibitem[Mengacci et~al., 2021]{10.3389/frobt.2021.713083}
Mengacci, R., Zambella, G., Grioli, G., Caporale, D., Catalano, M.~G., and
  Bicchi, A. (2021).
\newblock An open-source ros-gazebo toolbox for simulating robots with
  compliant actuators.
\newblock {\em Frontiers in Robotics and AI}, 8.

\bibitem[Venkatesh et~al., 2023]{venkatesh2023connected}
Venkatesh, N., Le, V.-A., Dave, A., and Malikopoulos, A.~A. (2023).
\newblock Connected and automated vehicles in mixed-traffic: Learning human
  driver behavior for effective on-ramp merging.
\newblock {\em arXiv preprint arXiv:2304.00397}.

\bibitem[Yalamanchi et~al., 2020]{9294553}
Yalamanchi, S., Huang, T.-K., Haynes, G.~C., and Djuric, N. (2020).
\newblock Long-term prediction of vehicle behavior using short-term
  uncertainty-aware trajectories and high-definition maps.
\newblock In {\em 2020 IEEE 23rd International Conference on Intelligent
  Transportation Systems (ITSC)}, pages 1--6.

\bibitem[Zhao and Malikopoulos, 2020]{zhao2020enhanced}
Zhao, L. and Malikopoulos, A.~A. (2020).
\newblock Enhanced mobility with connectivity and automation: A review of
  shared autonomous vehicle systems.
\newblock {\em IEEE Intelligent Transportation Systems Magazine},
  14(1):87--102.

\bibitem[Zhu et~al., 2019]{zhu2019probabilistic}
Zhu, J., Qin, S., Wang, W., and Zhao, D. (2019).
\newblock Probabilistic trajectory prediction for autonomous vehicles with
  attentive recurrent neural process.
\newblock {\em arXiv preprint arXiv:1910.08102}.

\end{thebibliography}

\end{document}